\documentclass[sigconf, nonacm]{acmart}
\usepackage{xcolor}
\usepackage{enumitem}
\usepackage{cleveref}
\usepackage{multirow}

\usepackage{listings}

\lstdefinestyle{pythonstyle}{
  language=C,
  basicstyle=\ttfamily\footnotesize,
  keywordstyle=\color{blue},
  commentstyle=\color{gray},
  stringstyle=\color{orange!90!black},
  showstringspaces=false,
  tabsize=2,
  frame=single,
  breaklines=true,
  linewidth=0.47\textwidth,
  numbers=left,                    
  numbersep=5pt,                   
  stepnumber=1                     
}

\AtBeginDocument{%
  }

\renewcommand\footnotetextcopyrightpermission[1]{}

\newcommand{\parhead}[1]{\vspace{0.2em}\noindent\textbf{#1}}
\newcommand{\ignore}[1]{}

\newcommand{\papername}{Selective-Index For Fast Compute of RAG Prefill by Exploiting Attention Invariance}
\newcommand{\shortname}{SIFT}
\newcommand{\Shortname}{\shortname\ }
\newcommand{\fullname}{\papername\ (\shortname)}

\begin{document}

\title{\shortname: \papername}

\author{Rya Sanovar}
\affiliation{%
  \institution{Georgia Institute of Technology}
  \city{Atlanta}
  \state{GA}
  \country{USA}
}

\author{Srikant Bharadwaj}
\affiliation{%
  \institution{Microsoft}
  \city{Redmond}
  \state{WA}
  \country{USA}
}

\author{Hritvik Taneja}
\affiliation{%
  \institution{Georgia Institute of Technology}
  \city{Atlanta}
  \state{GA}
  \country{USA}
}

\author{Moinuddin Qureshi}
\affiliation{%
  \institution{Georgia Institute of Technology}
  \city{Atlanta}
  \state{GA}
  \country{USA}
}

\begin{abstract}

Retrieval-Augmented Generation (RAG) injects Large Language Model (LLM) queries with relevant documents to improve response quality. This injection increases prompt length $(L)$ and slows \textit{time to first token (TTFT)} due to $O(L^2)$ attention. Unlike standard queries, RAG queries have a unique property of \textit{context reuse} where the same documents appear repeatedly across user queries. Thus, \textit{fully recomputing} documents for every RAG query does redundant compute and increases TTFT. Prior works precompute KV tensors of RAG documents offline and coarsely recomputing some tokens during online prefill. However, such KV reuse is often slower than full recomputation on modern GPUs due to high-latency disk transfers. Further, such a coarse-grained recomputation degrades accuracy. For example, CacheBlend degrades Llama-8B model's LongBench accuracy by $68\%$ and is just as slow as full recompute.

To address these limitations, this paper proposes \textit{\fullname}. SIFT processes documents offline and extracts fine-grained locations of high attention scores for each document, attention head, and model layer. Next, we identify the following \textit{attention invariance} insights that enable us to exploit the extracted locations during runtime: (1) \textit{Local-Attention Invariance}: The \textit{location} of high attention scores within a document remains invariant to surrounding documents. This helps us predict the location of high scores where the document attends to itself. (2) \textit{Cross-Attention Consistency}: Keys with high intra-document attention also attract cross-attention from subsequent documents. This helps us predict the location of high scores where the document attends to future documents. Critically, \textit{\Shortname stores no KV data and only stores locations of high attention scores} in the form of two compact bit vectors. \shortname's storage is only a few KB of bit vectors, up to 24,000$\times$ smaller than KV tensors and obviating costly disk transfers. During the query processing (prefill), SIFT uses a custom attention kernel to read the bit vectors and computes the attention only for the marked locations. SIFT improves TTFT by up to $1.71\times$ while holding the average accuracy within $1\%$ of full recompute.

\end{abstract}

\maketitle

\section{Introduction}

\begin{figure*}[htb!]
    \centering
    \includegraphics[width=\textwidth]{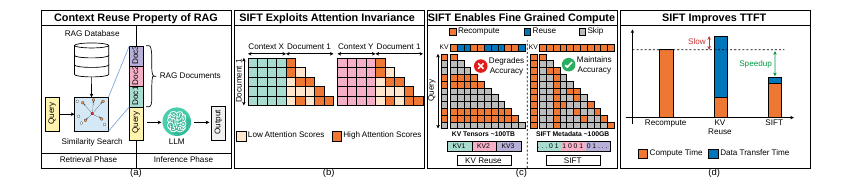}
    \caption{(a) RAG documents make up a large portion of the input prompt. (b) \Shortname exploits attention invariance to locate high attention scores offline. (c) \Shortname performs fine-grained compute and maintains accuracy compared to KV reuse methods. (d) Retrieving KV tensors from disk is slower than full recompute. SIFT's metadata is 24,000x smaller and can be stored in  DRAM, enabling efficient compute.}
    \label{fig:intro}
     \vspace{-1em}
\end{figure*}

Retrieval-Augmented Generation (RAG) has emerged as a popular augmentation to modern Large Language Model (LLM) capabilities~\cite{rag1, rag2, rag3, rag4, rag5}. RAG enhances the ability of the model to generate accurate and contextually relevant responses and obviates costly model re-training~\cite{wampler}. A canonical RAG pipeline consists of two phases: (1) the retrieval phase that finds top-k relevant documents by conducting a similarity search against the user's query and the RAG database~\cite{ivf2, ivf}, and (2) the generation phase, where the retrieved documents are prepended to the user's query to form an expanded input prompt to the LLM (Figure~\ref{fig:intro}(a)). 

However, prepending RAG documents to the user query increases the context length of the prompt and leads to increase in time to first token (TTFT). TTFT is the time taken during the prefill phase of inference to generate the first output token. A high TTFT not only degrades user experience in AI applications but also lowers throughput leading to smaller batches in the following decode phase. Long context lengths $L$ increase TTFT primarily due to the compute-heavy prefill-attention layer that scales as $O(L^2)$. For instance, attention takes up 47\% of TTFT at just 64K context length for MiniMax M2.5 MoE. Ideally, we would want to reduce time spent in attention to improve TTFT.

Unlike typical long context prompts, RAG-injected prompts have a defining characteristic of \textbf{context reuse}. RAG documents frequently repeat across different user queries in different orderings and combinations~\cite{vectorliterag}, leading to significant overlap in input prompts to the LLM. Therefore, this prevalence of context reuse can be exploited to reduce prefill compute and improve TTFT.

If this context reuse property is not exploited at all, the model treats every new RAG-injected query as completely novel and \textit{fully recomputes} it every time, leading to increased TTFT for every new user query. In order to exploit this context reuse property, Key/Value (KV) tensors of RAG documents can be precomputed offline and re-used during online inference~\cite{promptcache}. Such \textit{Full KV reuse} greatly reduces prefill compute load by entirely bypassing compute for the RAG portion of the prompt. However, full KV reuse doesn't account for the contextual interactions (cross-attention) across documents leading to severe accuracy degradation. 


Recent works~\cite{blend, epic} tried to reduce this accuracy gap by recomputing the KVs of a few selected tokens. For instance, CacheBlend recomputes ${\sim}$20\% of document tokens across all model layers. Although this improves accuracy over full KV reuse, it still falls short of full recompute's accuracy. For instance, CacheBlend~\cite{blend} degrades Llama 8B accuracy on LongBench tasks by $68.25\%$ (\Cref{fig:llama3-benchmark}).

The root cause for this accuracy degradation is its coarse-grained recomputation that recomputes the same token set across all model layers.
Recomputing KVs of a fixed token set refreshes attention scores at only fixed locations in the attention layer. However, attention score patterns are well-known~\cite{minference, sampleattn, FlexPrefill, sparseattn} to be diverse for every (context, attention head, model layer) combination, and correctly recovering attention scores is necessary to maintain accuracy.  

Worse still, reading stale KV tensors from storage is slower than recomputing them on modern GPUs. On NVIDIA H200s, we find that CacheBlend is markedly slower than full recompute at a wide range of context lengths (\Cref{fig:all-models-benchmark}). Thus, even when KV-reuse methods reduce compute load, the speedup is diluted, and often eliminated, by the latency of transferring the large KV cache itself. 

The \textbf{goal} of this work is to improve the TTFT of RAG workloads over full recompute while maintaining accuracy. To achieve this, we must perform only the minimum amount of computation necessary to recover accuracy while significantly reducing the compute load of prefill.

To maintain accuracy and reduce compute, only high attention scores in RAG prefill need to be recovered correctly since high scores contribute most to the attention output which determines the final response. To that end, we propose \textit{\fullname}, which encodes the predicted locations of high attention scores, as shown in Figure~\ref{fig:intro}(c). SIFT is based on two key insights.  

Our first insight is that, because documents recur across queries, SIFT can analyze RAG documents offline to identify key properties.  For example, this analysis can help SIFT materialize the true attention matrix for each (document, head, layer) and directly observe the nature of the attention score distributions within that document. 

RAG documents co-occur with other documents in compositions unknown until retrieval time. In the actual online query, documents attend not only to themselves (local-attention) but also to one another (cross-attention). Therefore, the \textit{values} of the actual attention scores during runtime are \textit{different} from the values observed offline, depending on the document mix in the RAG query. 


Our second insight is to establish the \textbf{attention invariance} properties of RAG documents (\Cref{fig:intro}(b)), enabling us to exploit offline attention scores for any runtime document composition. First, \textbf{local-attention invariance:} we observe that the \textit{location} of high attention scores within a document's self-attention tend to remain invariant to surrounding documents. Second, \textbf{cross-attention consistency:} We observe that keys in a document that attract high attention scores within that document tend to also attract strong cross-attention from future co-resident documents. 

These two attention invariance properties jointly enable SIFT to accurately predict the \textit{locations} of high local attention and cross attention scores without knowing the actual composition of documents. SIFT encodes the predicted locations of high attention scores as two compact bit vectors per document. At prefill time, SIFT uses a custom sparse attention kernel to read \shortname's bit vectors and computes the attention score \textit{only} at the marked locations. 

 
\Shortname does not store any KV data and only stores the locations of high attention scores. This reduces per-document storage from MBs-GBs of KV tensors to only a few KBs of location metadata, eliminating the storage$\rightarrow$GPU transfer bottleneck that crippled prior KV-reuse schemes. Per Figure~\ref{fig:intro}(d), SIFT improves TTFT by up to $1.71\times$ over full recompute while having accuracy within $1\%$ of full recompute.

Overall, this paper makes the following contributions:
\begin{enumerate}[noitemsep]
\vspace{-0.05 in}

    \item We identify \textbf{local-attention invariance}: the attention sparsity pattern of a document's self-attention is invariant to its surrounding context, enabling prediction of local-attention.
    \item We identify \textbf{cross-attention consistency}: Keys that are highly attended to within a document will be attended to by future documents, enabling prediction of cross-attention.
    \item We develop \textbf{\shortname}, which encodes the locations of high local and cross attention scores into 2 compact bit vectors and achieves up to 24,000$\times$ size reduction over KV cache, delivers up to 1.71$\times$ TTFT speedup and maintains average accuracy within 1\% of full recompute.
\end{enumerate}

\section{Background and Motivation}

\subsection{Prefill Phase of LLM Inference}
\label{sec:prefill}

Auto-regressive LLM inference consists of two computationally distinct phases. The \emph{prefill phase} consumes the entire prompt of $L$ tokens in parallel and produces the first output token. The latency of prefill phase is termed as \emph{time to first token (TTFT)}. A fast TTFT is necessary as responsiveness is key to user satisfaction. Moreover, fast TTFT is essential in order to finish prefill phase faster so that more requests can be batched together in the following memory-bound \emph{decode phase} of inference.
Prefill also generates the Key/Value (KV) Cache of all prompt tokens, which is re-used in the subsequent \emph{decode phase} where tokens are generated auto-regressively, each attending to the past KV cache. 

\subsection{RAG-injected Inference}
\label{sec:raginference}
\begin{figure}[htb!]
    \centering
    \vspace{0.1 in}
    \includegraphics[width=1\linewidth]{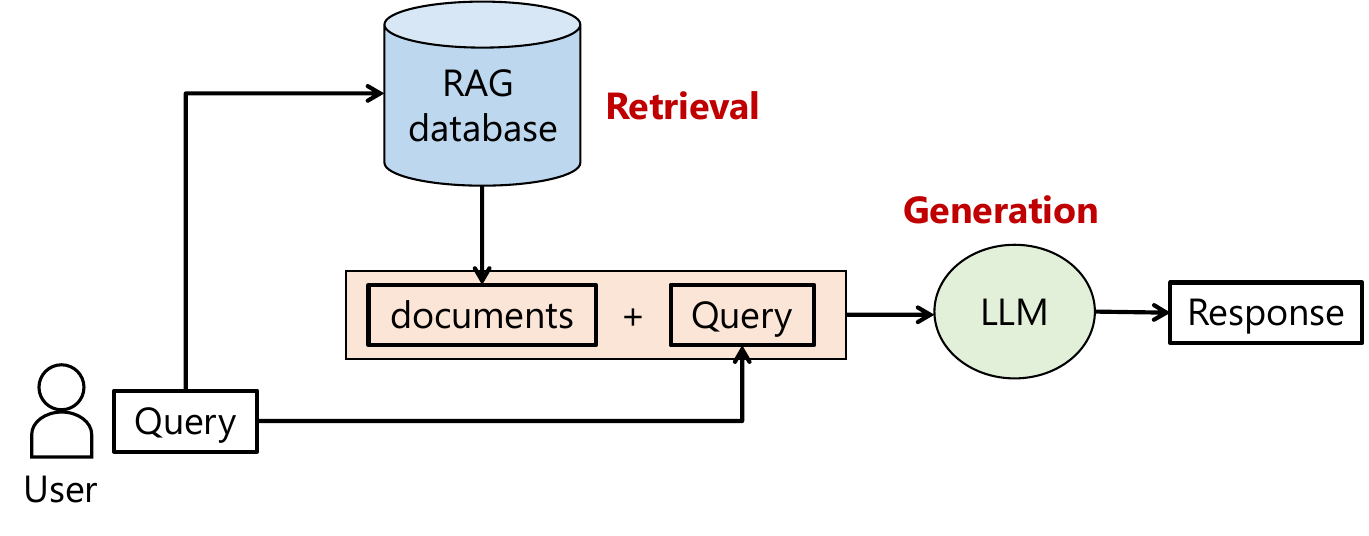}
    \caption{A RAG pipeline consists of (1) a document retrieval phase followed by (2) response generation phase.}
    \label{fig:rag-pipeline}
\end{figure}

Retrieval Augmented Generation (RAG) is a post-training augmentation on LLMs where relevant documents are retrieved and prepended to the original user query. This inflation of the original query with useful documents not only helps the model generate high quality and contextually relevant responses but also obviates costly model re-training. A RAG pipeline consists of two phases: (1) the retrieval phase and (2) the generation phase~(see \Cref{fig:rag-pipeline}). 
The retrieval phase takes the user query as input and performs an extensive similarity search~\cite{ivf, spann, diskann} over the vector database containing millions to billions of document embeddings. In the generation phase, the RAG prepended query forms the prompt for the prefill phase of inference.

\subsection{Overhead of Attention during Prefill}
Prepending the originally small user query with RAG documents significantly increases the context length $L$ of the input prompt to the LLM. During the prefill phase, the compute for MLP and attention projections scales as $O(L)$ whereas the compute in the self-attention operation scales as $O(L^2)$. Therefore, as context lengths increase, the proportion of time spent in attention increases significantly compared to the other operations in prefill.

\begin{figure}[htb!]
    \centering
    \includegraphics[width=\linewidth]{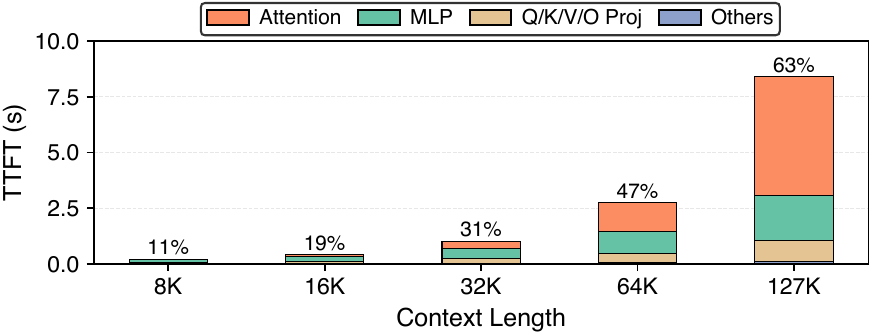}
    \caption{Breakdown of TTFT for MiniMax M2.5 on 4 H200s: Time spent in attention increases significantly~(grows with $O(L^2)$) with context length~($L$).}
    \label{fig:ttft_breakdown}
\end{figure}

\Cref{fig:ttft_breakdown} shows the breakdown of time spent during different operations of the MiniMax M2.5 MoE model. We observe that as we increase the context length from 8K to 127K, the time spent in attention increases from 11\% to 63\%. So, in this work, we focus on optimizing the time spent in attention for RAG workloads.

\subsection{Context Reuse Property of RAG}

RAG workloads offer a unique opportunity to optimize prefill because retrieved documents are drawn from a fixed corpus known ahead of time. 
Moreover, it has been observed that across different users and queries, the same documents are frequently retrieved~\cite{vectorliterag, telerag}.

This offers a unique advantage where RAG documents can be preprocessed offline to extract useful information that can be leveraged to reduce the compute load of the prefill of an online RAG query. Since these documents are reused across multiple queries, the cost of offline preprocessing for a RAG corpus can be amortized over many online queries.

\begin{figure}[htb!]
    \centering

    \includegraphics[width=0.49\linewidth]{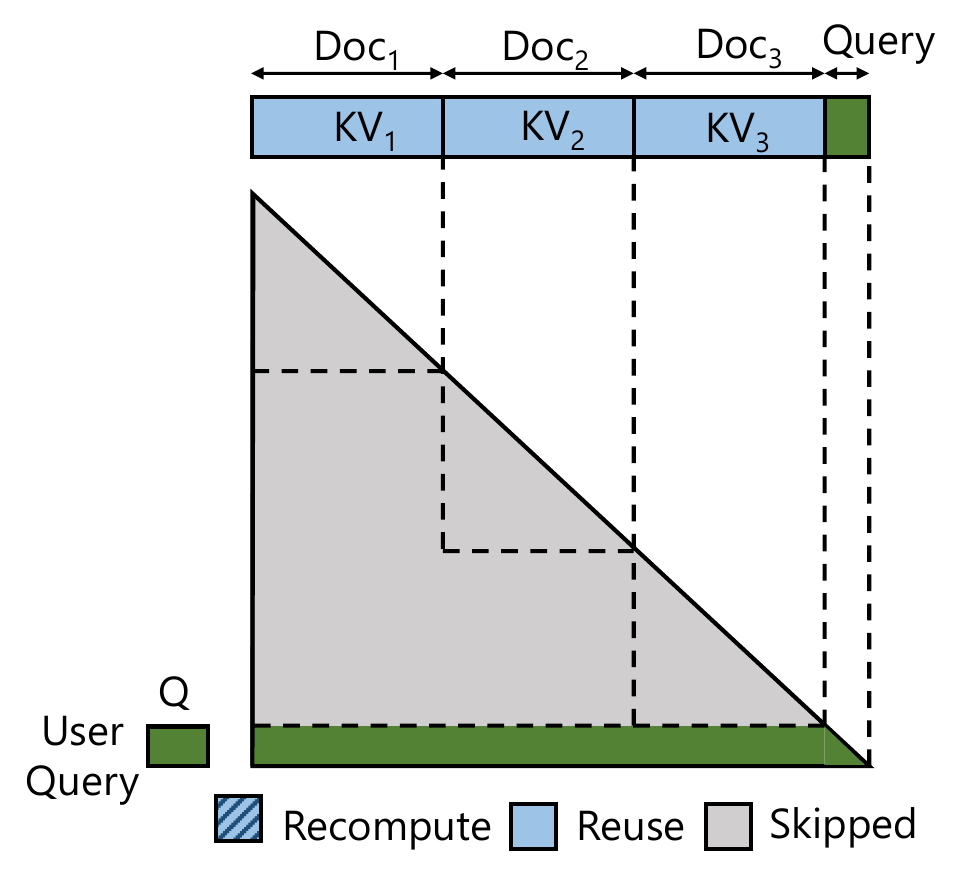}
    \includegraphics[width=0.49\linewidth]{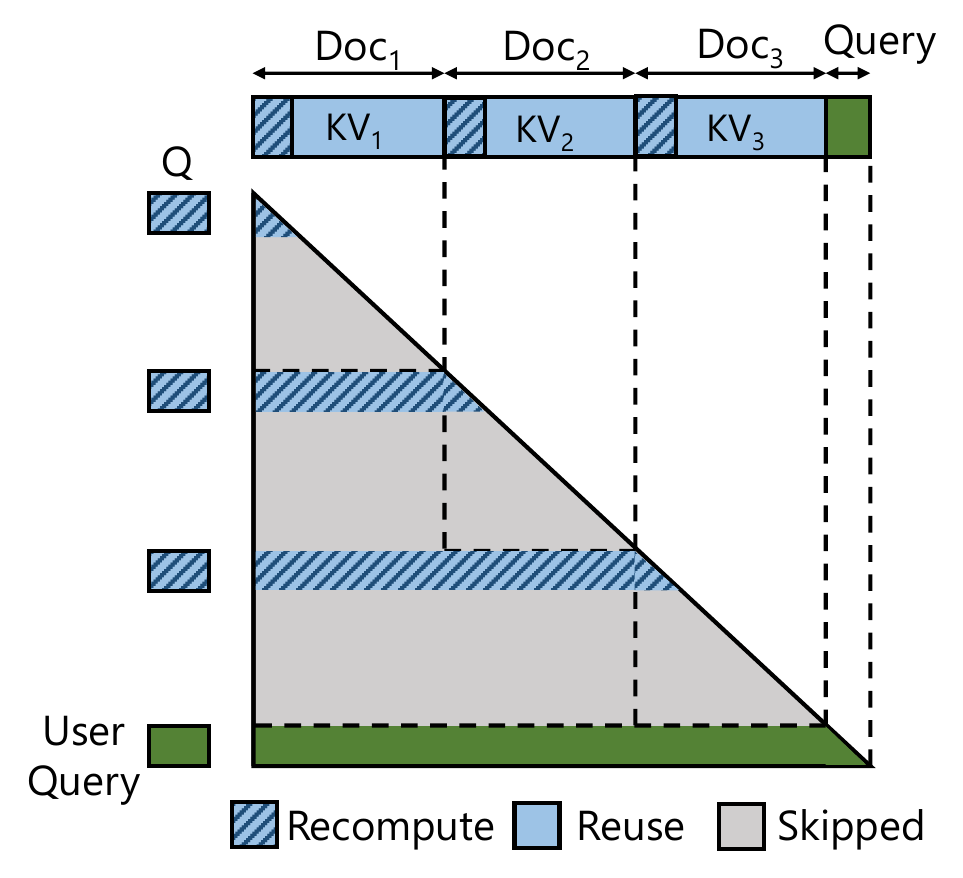}
    \caption{(a) Full KV Reuse: skips cross-attention, provides poor accuracy (b) KV Reuse with Selective Recompute: Coarse-grained selective recompute degrades accuracy.}
    \label{fig:cacheblend-pictorial}
 \vspace{-0.15 in}
\end{figure}

\subsection{Accelerating Prefill via KV Reuse}
\label{sec:prior-works}

\ignore{
\begin{figure}[htb!]
    \centering
    \includegraphics[width=0.7\linewidth]{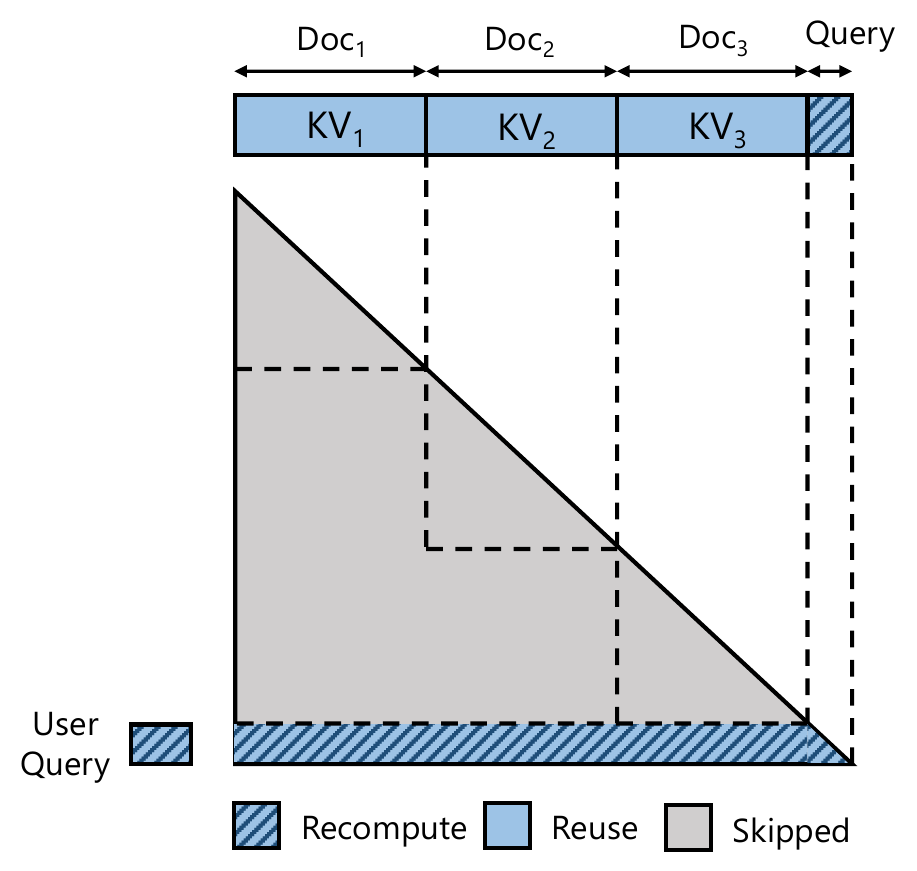}
    \caption{Full KV Reuse: Document KVs are reused which skips cross-attention across documents, resulting in severe accuracy degradation.}
    \label{fig:kv-reuse}
\end{figure}
}

\begin{figure*}[htb!]
    \centering
    \includegraphics[width=\textwidth]{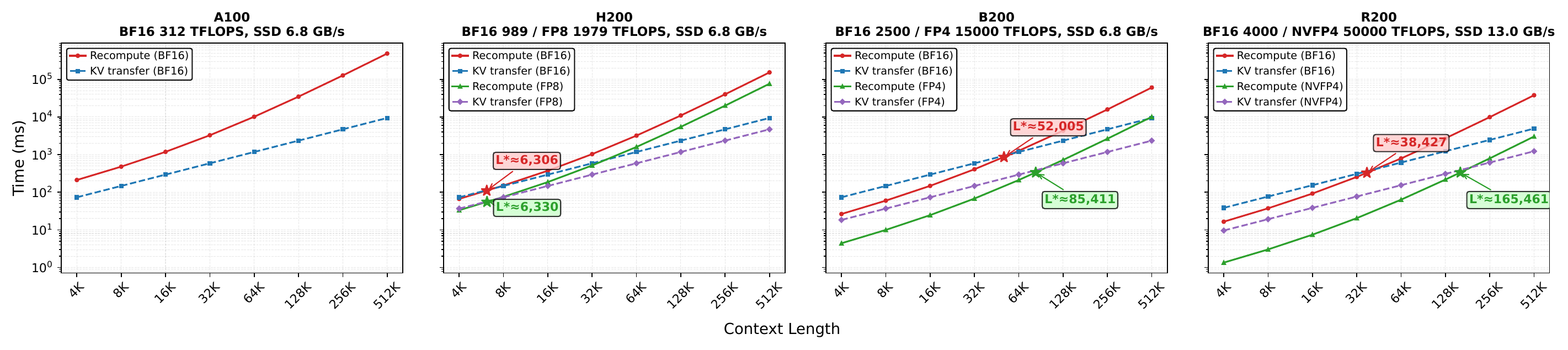}
    \caption{Full Recompute and KV transfer time for different generations of DGX systems for a Llama 8B-like model architecture. The window of context lengths at which full recompute is faster than transfering KVs from disk expands for every new GPU generation especially with better compute support for emerging datatypes.}
    \label{fig:crossover-in-dgx}
\end{figure*}

\parhead{Full KV Reuse~\cite{promptcache}.}
This approach naively reuses the KVs of RAG documents and only recomputes the short user query, as shown in \Cref{fig:cacheblend-pictorial}(a). This significantly reduces prefill compute load by effectively omitting repeated $O(L^2)$ attention of RAG documents.  As the KV cache of each document was computed independently offline, it captures only local-attention of each document with itself. As a result, full KV reuse fails to capture cross-attention across documents, leading to significant accuracy degradation.

\parhead{KV Reuse with Selective Recompute~\cite{blend, epic, fusionrag}.}
To maintain accuracy, prior works employ \emph{selective recomputation} of a subset of document tokens, as shown in \Cref{fig:cacheblend-pictorial}(b). For example,  CacheBlend~\cite{blend} compares fully recomputed KVs with the stale precomputed KVs in the first model layer and selects the top ${\sim}$20\% of tokens with the highest attention deviation. Only these tokens are recomputed across all subsequent layers, whereas stale KVs are reused for the rest. 


\subsection{Limitation of Prior Works}
\label{sec:tradeoff}

KV-reuse based solutions face three key challenges: 


\parhead{Accuracy.} CacheBlend has accuracy degradation that is as high as $73\%$ on certain tasks (see \Cref{fig:all-models-benchmark}, HotpotQA for Llama 8B). This is because of its coarse-grained recomputation of the attention layer. The same token set is recomputed across all layers and heads and the ${\sim}$20\% budget is not adaptive to each document's unique attention score distribution, as shown in ~\Cref{fig:cacheblend-pictorial}(b).

In fact, KV reuse \textit{necessitates} coarse-grained recomputation of a fixed token set across all model layers. This is because layer $\ell+1$'s KVs are produced from layer $\ell$'s attention output, so recomputing a different token set at layer $\ell$ changes layer $\ell+1$'s KVs and invalidates the precomputed KVs that would be reused otherwise. Enabling KV reuse therefore requires recomputing the same token set at every layer, so that the non-recomputed tokens' KVs can be reused.

In contrast, attention score distributions are unique for every (context, attention head, model layer) combination~\cite{minference, sparseattn, FlexPrefill, sampleattn}. Therefore, the coarse-grained recompute of KV reuse methods fails to capture the true diverse attention distribution of the RAG-injected query. The focus of this work is to develop a fine-grained recompute strategy that recovers the diverse attention score patterns of every (document, attention head, model layer).

\parhead{Memory Capacity.}
For a model with $L_{\text{layers}}$ layers, $K$ KV heads, and head dimension $d$, the KV cache of a single token requires $2 \cdot L_{\text{layers}} \cdot K \cdot d \cdot 2$ bytes (in BF16). For Llama 3.1 8B, this is 128\,KB per token. A typical WikiAll RAG database hosts about 88M passages, and each passage has about 128 tokens. According to RAG retrieval studies~\cite{vectorliterag, telerag}, the top 20\% of clusters account for 60--90\% of accesses. Therefore, only persisting the KV cache of the top 20\% clusters requires $\sim\!268$\,TB of storage. This is orders of magnitude beyond the DRAM capacity of modern servers. Therefore, KV caches of RAG documents must reside on larger, but slower disks. 

\parhead{Latency.} Reading KV cache from disk can be either faster or slower than recompute, depending on the compute capacity and the storage bandwidth. KV transfer time scales as $O(L)$, while prefill recompute scales as $O(L^2)$. However, as \Cref{fig:crossover-in-dgx} shows, GPU compute has grown ${\sim}12.8\times$ while NVMe SSD bandwidth has grown only ${\sim}1.9\times$. This asymmetry in compute and memory transfer capabilities results in model-hardware configurations where reading KVs from disk becomes slower than full recompute.


For example, on a DGX B200, recompute beats disk transfer for any context length below 52K tokens. Therefore, KV-reuse methods start to become disk-bound and waste bandwidth on transferring KVs that the GPU could have regenerated faster. Furthermore, each new generation brings native low-precision compute (FP8 on H200, FP4 on B200, NVFP4 on R200) while SSD bandwidth stagnates. Thus, the window of context lengths where GPU compute is faster expands with every generation, making KV transfer the bottleneck. 


\subsection{Goal Of This Paper} 
The goal of this work is to improve the TTFT of RAG prefill on modern GPUs while maintaining the accuracy of full recompute. To accomplish this, we need a fine-grained recompute strategy that performs minimal computation to maintain accuracy while significantly reducing the compute load of RAG prefill. We next show that the context reuse property of RAG and {\em attention-invariance} can enable such fine-grained recomputation with high accuracy.

\section{Insight for Minimal Recomputation}
\label{sec:minimal-state-representation}


To minimize computation while maintaining accuracy, only the high-attention scores in the RAG-injected query's attention layers need to be recovered correctly. The attention output is the weighted sum of the value vectors, with weights given by the attention scores. High attention scores contribute the most to the attention output. Therefore, accurate recovery of these scores can maintain accuracy, and low scores can be skipped to reduce computation.

To this end, we first leverage the context reuse property of RAG and make RAG documents go through prefill independently offline. This reveals the nature of attention score distributions in the self-attention layer of each document.

However, during runtime, RAG documents co-occur with other documents in compositions unknown until after the retrieval phase. In the online RAG-injected query, documents attend to themselves (local-attention) and to each other (cross-attention). Thus, the \textit{values} of the actual attention scores during runtime prefill are \textit{different} from the values of attention scores observed in offline prefill.

In order to exploit the attention score distribution knowledge acquired offline under any runtime document composition, we identify two \textbf{attention invariance} properties of RAG documents in the following sections.

\subsection{Decomposing the RAG Attention Matrix}
\label{sec:decomposing-attention-matrix}
\begin{figure}[htb!]
\centering
\includegraphics[width=0.45\linewidth]{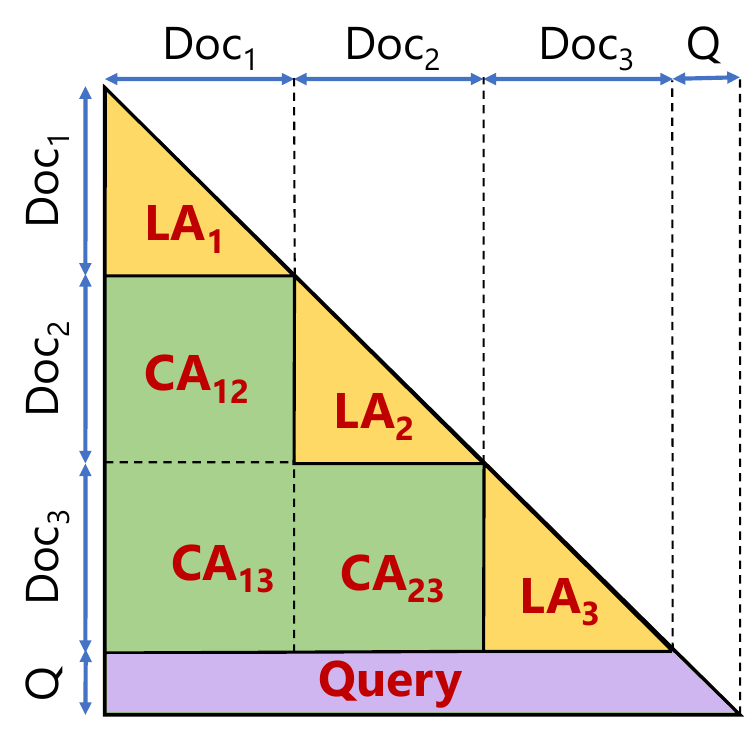}
\caption{RAG's prefill-attention matrix can be decomposed into cross-attention (CA), local-attention (LA) and the query.}
\label{fig:decomp-attn-matrix}
\end{figure}

The attention matrix in RAG Prefill can be divided into 3 submatrices (\Cref{fig:decomp-attn-matrix}). The first is the \textbf{Local-Attention (LA)} submatrix, which captures the self-attention patterns within a document. The second is the \textbf{Cross-Attention (CA)} submatrix, which captures the attention patterns across different documents. The third is the \textbf{Query} submatrix, which captures the attention patterns between the user query and all the past tokens (query and RAG documents). The query submatrix is always recomputed because it is the novel portion. 

Predicting high attention scores for a RAG portion of the prompt requires solving two distinct problems: (1) predicting high attention scores within a document (local-attention), and (2) predicting high attention scores across documents (cross-attention). We address each with a separate insight.

\subsection{Insight 1: Local-Attention Invariance}
\label{sec:local-attention-invariance}

\begin{figure}[htb!]
\centering
\includegraphics[width=\linewidth]{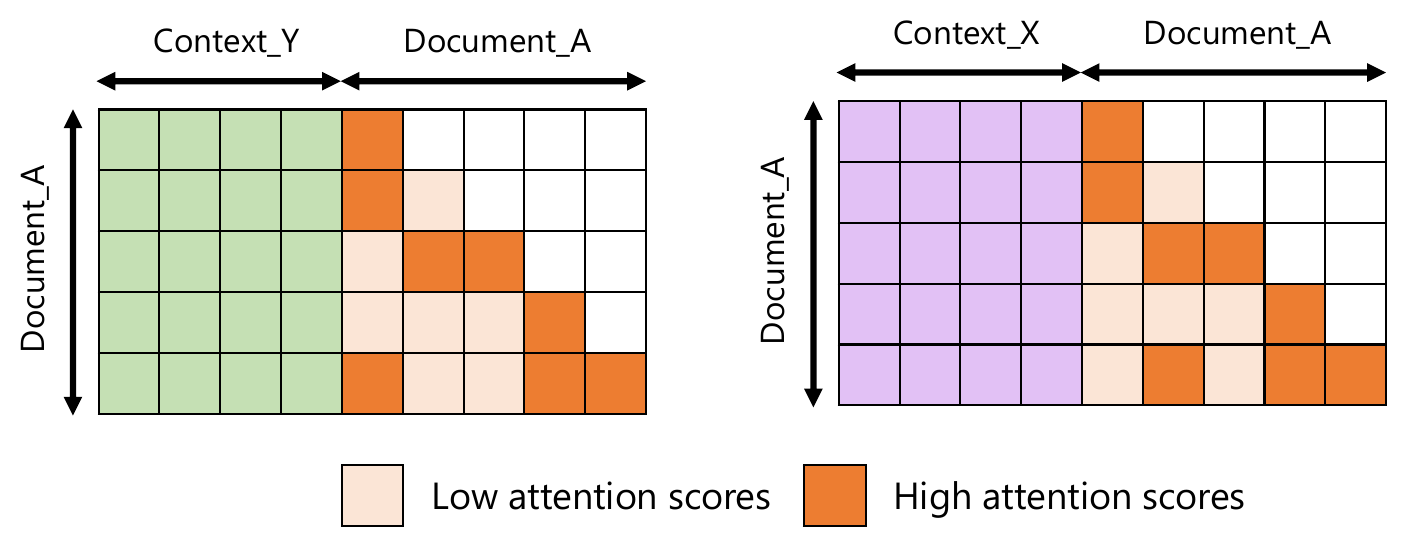}
\caption{Locations of high local attention scores in a document are invariant of prepended context.}
\vspace{-0.15 in}
\label{fig:la-pictorial}
\end{figure}

We observe that the \textit{locations} of high attention scores in the local attention submatrix of a document are invariant to the changes in the prepended context, in particular, which other documents appear with the given document. When the prepended context changes, the \emph{values} of attention scores within the local attention submatrix change, but the \emph{locations} of high attention scores remains stable regardless of the prepended context (\Cref{fig:la-pictorial}). Intuitively, the correlation between tokens in a document varies with the global context, strengthening or weakening depending on the context, but the \emph{gradient} of this correlation remains invariant.

\subsection{Validation of Local-Attention Invariance}

\begin{figure}[htb!]
\centering

\includegraphics[width=\linewidth]{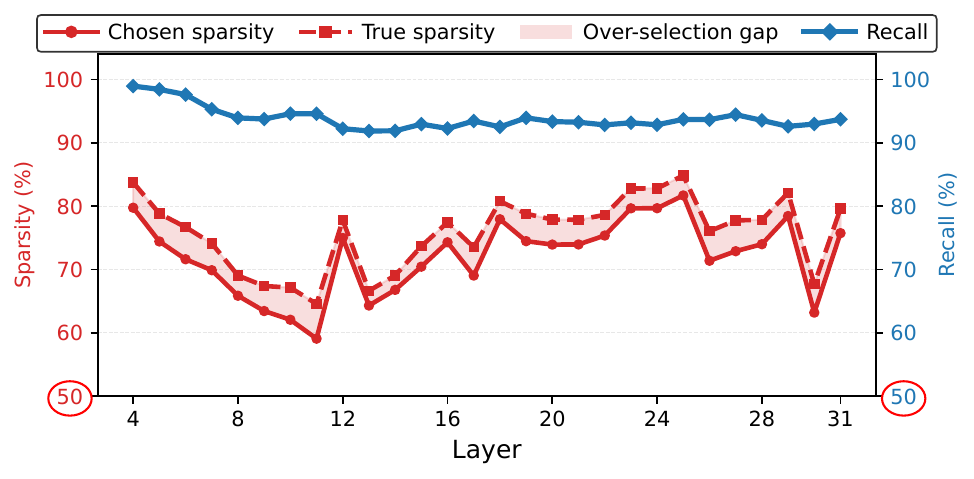}
\vspace{-0.15 in}
\caption{True local-attention sparsity (\%) and sparsity chosen by our offline analysis. We correctly identify $93.9\%$ (recall) of high local-attention score locations.}
\vspace{-0.2 in}
\label{fig:la-evidence}
\end{figure}

To quantify local-attention invariance, we record the locations of high attention scores in a document when it is processed offline versus in its local-attention submatrix when it is processed with a prepended context. We select any attention score ${>=}\,0.001$ as a high local attention score for this study. We find that $93.89\%$ of the locations of truly high local-attention scores are also along the same locations recorded from the offline standalone pass (recall in \Cref{fig:la-evidence}). Our prediction also conservatively over-selects attention scores: a local attention sparsity ratio of $72.7\%$ was chosen, compared to a slightly higher ground truth sparsity of $76.4\%$.

\subsection{Insight 2: Cross-Attention Consistency}
\label{sec:cross-attention-consistency}

\begin{figure}[htb!]
\centering
\includegraphics[width=0.7\linewidth]{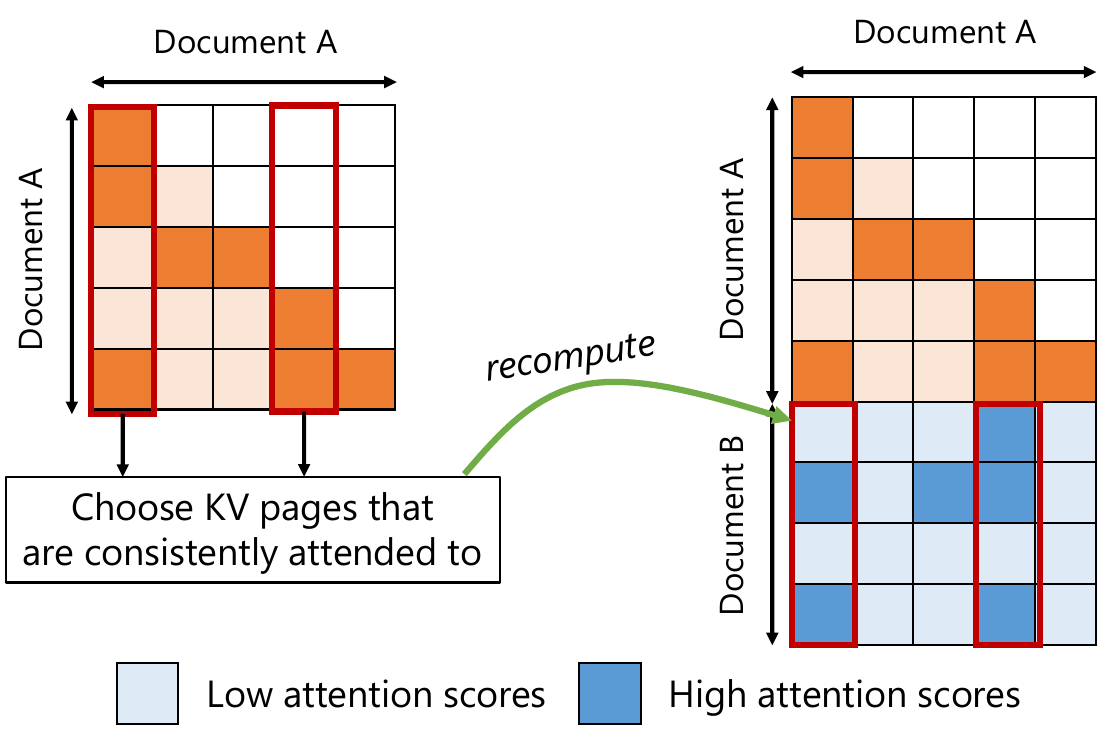}
\caption{KV tokens that accrue consistent high cross-attention scores within a document also accrue high cross-attention scores from future documents.}
\vspace{-0.2 in}
\label{fig:ca-pictorial}
\end{figure}

We observe that Keys which accrue high attention scores during the standalone prefill of a document also tend to be attended to strongly by tokens from future documents (\Cref{fig:ca-pictorial}). We call this property {\em Cross-Attention Consistency}. Intuitively, a Key token that consistently attends strongly to many of the query tokens within a document usually encodes semantically important content. Thus, tokens from other documents will also attend to the same important content.

\subsection{Validation of Cross-Attention Consistency}

\begin{figure}[htb!]
\centering
\includegraphics[width=\linewidth]{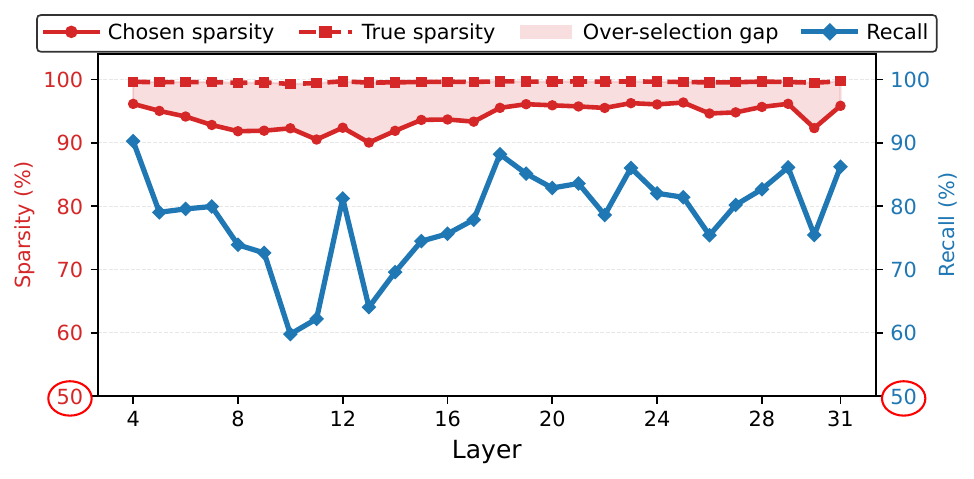}
\vspace{-0.25 in}
\caption{Cross-attention sparsity (\%) -- $80.1\%$ (recall) of high attention score locations were correctly predicted.}
\vspace{-0.2 in}
\label{fig:ca-evidence}
\end{figure}

To quantify cross-attention consistency, we record the Key tokens with the highest concentration of high attention scores within a document when it was processed offline, and also record the locations of actual high attention scores in the cross-attention submatrix of that document against future documents. We consider any attention score ${>=}\,0.01$ as a high attention score and select Key tokens with ${>}10\%$ concentration of high scores along them. We conducted this study over 50 samples of LongBench~\cite{longbench} for Llama 8B. We find that ${80.12\%}$ of the locations of truly high cross-attention scores are present in the offline pass (recall in \Cref{fig:ca-evidence}). Our analysis chose cross-attention sparsity of $94.2\%$, compared to a higher ground truth cross-attention sparsity of $99.6\%$. This over-selection is natural, since we recompute entire key columns rather than just individual (query, key) score cells.

\section{\Shortname Design}
\label{sec:design}



Based on the key insight of attention-invariance, we propose  \textit{\fullname}. SIFT reduces the computational cost of RAG prefill while maintaining high accuracy.  SIFT encodes the locations of high-attention scores detected by the analysis in \Cref{sec:minimal-state-representation} as metadata that takes up minimal space. SIFT contains two key encodings: the \textbf{Local-Attention (LA) bit vector} and the \textbf{Cross-Attention (CA) bit vector}, which together encode the locations of high attention scores in the RAG prefill attention matrix.

\subsection{Local-Attention Encoding}
\label{sec:la-encoding}

Local-Attention Invariance (\Cref{sec:local-attention-invariance}) informs us on the locations of high attention scores within the local attention submatrix. We need a compact encoding that captures these locations per (document, head, layer).

The Local-Attention (LA) bit vector is designed to achieve this. To create this bit vector, we firstly abstract sparsity at a tile-size T (group of $(T \times T)$ tokens) granularity. We then tile the lower triangular attention matrix (due to causality) and view it as a grid of $(T \times T)$ tiles. A tile is marked for recomputation by setting its corresponding bit to 1, if it contains atleast 1 attention score above a pre-defined threshold $\alpha\,(=0.001)$, otherwise the corresponding bit is set to 0.

Then, these boolean bits are packed MSB-first, enumerated row-major in the lower triangle: (0,0), (1,0), (1,1), (2,0) ... and so on. Each document's bits are byte-aligned (padded to next byte boundary). Tiles set to 1 are recomputed during online RAG prefill, while tiles set to 0 are skipped. 



\subsection{Cross Attention Encoding}
\label{sec:ca-encoding}

Cross-Attention Consistency (\Cref{sec:cross-attention-consistency}) informs us on the locations of high attention scores in the cross-attention submatrix. 

Similar to local-attention encoding, we abstract cross-attention at a tile/page of width T granularity. For each KV page, we compute the fraction of tokens within that page that have attention scores greater than a pre-defined threshold $\beta\,(=0.01)$. If that fraction exceeds another pre-defined threshold $\gamma\,(=10\%)$, we set the corresponding bit for that KV page to 1, indicating that this page must be recomputed for future documents. Otherwise, the bit is set to 0.
KV pages to be computed or skipped are encoded in this manner for every (document, head, layer).


\subsection{\Shortname Storage}
\label{sec:ic-storage}

\Shortname stores only two bit vectors per (layer, head, document) tuple. Given a tile/page size $T$, the number of KV pages in a document of length $L$ is $P = \lceil L/T \rceil$. Therefore, the size of the LA bit vector is $\lceil P(P+1) / (2 \cdot 8) \rceil$ bytes and the size of the CA bit vector is $\lceil P / 8 \rceil$ bytes. Importantly, while \Shortname scales quadratically in number of pages as $O(P^2)$, each unit of storage is a \emph{single bit}, compared to the KB-scale KV vectors. For example, for MiniMax-M2.5 at $64K$ context length, and assuming $P \approx 8$ pages per document, the LA bit vector stores $36$ bits and the CA bit vector stores $8$ bits per (head, layer, document). Aggregated across all documents, heads and layers, \Shortname metadata is  0.98\,MB of which 0.79\,MB is LA bit vectors and 0.17\,MB is CA bit vectors. Therefore, \Shortname is approximately 20,000$\times$ smaller than storing the 15.1\,GB KV Cache for the same context. 

\begin{figure}[htb!]
\centering
\includegraphics[width=\linewidth]{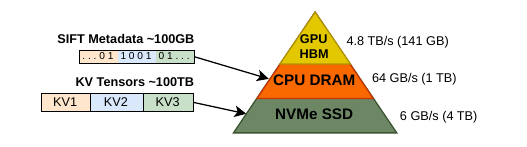}
\caption{Storage sizes of SIFT and KV Reuse Methods for a typical RAG database: SIFT's metadata is about 3 orders of magnitude smaller and can easily reside in faster DRAM.}
\label{fig:storage-pictorial}
\end{figure}

For the same WikiAll database example discussed in \Cref{sec:tradeoff}, persisting \Shortname metadata for the top 20\% RAG clusters on CPU DRAM would require 101 GB (assume $T=16$) compared to 268 TB of KV Cache, a 2,700$\times$ reduction that makes \Shortname storage feasible on faster CPU DRAM (\Cref{fig:storage-pictorial}), eliminating the disk bottleneck that made KV-reuse methods counterproductive to performance. \Cref{sec:storage-scaling} further details the storage scaling of \Shortname vs KV tensors across diverse model architectures. 

\subsection{End-to-End System Design}
\label{sec:e2e-pipeline}

\begin{figure}[htb!]
\centering
\includegraphics[width=\linewidth]{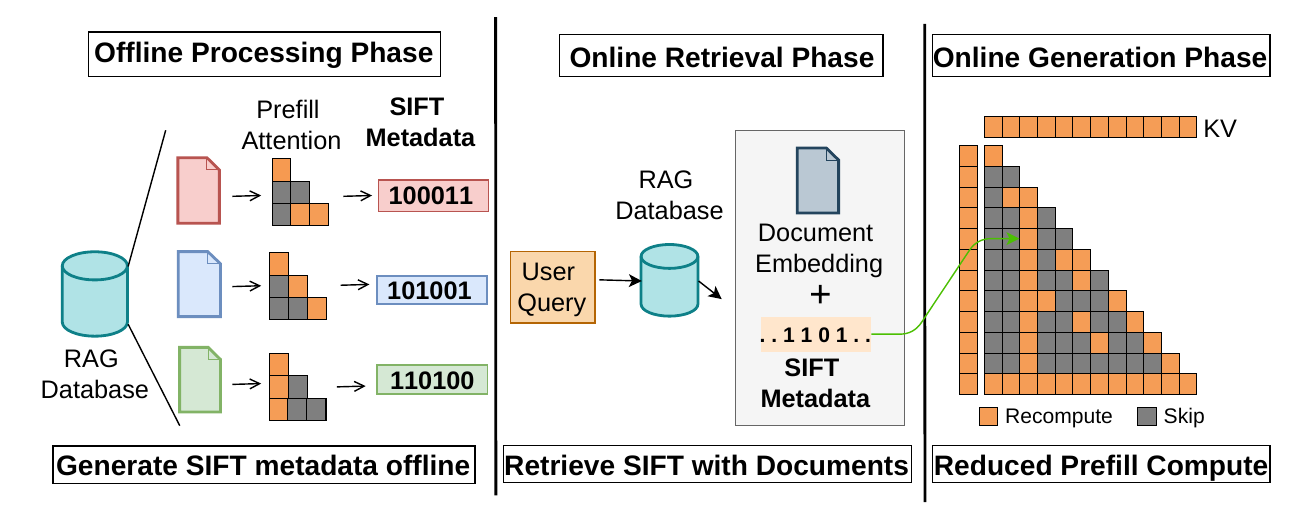}
\caption{\Shortname Operation: (a) Metadata is extracted offline through dense prefill. (b) \Shortname metadata is retrieved along with top-K documents. (c) Selective prefill is performed on locations identified by the \Shortname metadata.}
\label{fig:sys-design}
\vspace{-0.2 in}
\end{figure}

\parhead{Generating \Shortname Metadata.} \Shortname is created offline by performing dense prefill on each RAG document. \Shortname metadata is stored in CPU DRAM, co-located with the vector embeddings of the document in the RAG database (\Cref{fig:sys-design}(a)).

\parhead{Query-Time Consumption.} During the RAG retrieval phase (\Cref{fig:sys-design}(b)), the SIFT metadata is retrieved along with top-k documents. Then, the online generation phase consumes the \Shortname bit vectors and performs fine-grained selective recompute during the prefill phase (\Cref{fig:sys-design}(c)).

\section{Implementation Details}
\label{sec:implementation}

We implement two custom kernels that together enable RAG prefill with \shortname: (1) a decoding kernel that decodes the bit vectors into explicit tile indices, and (2) a custom sparse attention kernel that consumes these indices to compute only the tiles at the given index locations.

\subsection{\Shortname Metadata Decoding Kernel}
\label{sec:decoding-kernel}
The \Shortname metadata decoding kernel generates a list of \texttt{sparse}- \texttt{\_n\_indices} per layer. This is a packed list of \texttt{int32} integers for every (head, M-block) (N-block is the Key's tiled column and M-block is the Query's tiled row).

The decoding kernel does two passes over \shortname's bit vectors. Pass~1 counts the number of tiles for both cross-attention and local-attention and Pass~2 uses this count to read bits from the bit vector and determine the actual KV indices a given (head, M-block) must attend to.

A CTA of 128 threads is launched per (head, M-block) tuple. In Pass~1, for document M-blocks, threads cooperatively scan the CA and LA bit vectors to count tiles marked 1 for that (head, M-block) tuple. For user query M-blocks, the count is the number of causal N-blocks, as it is fully recomputed. 

In Pass~2, each threadblock now uses the counts generated in Pass~1 to figure out the range of bits in the CA and LA bit vector that belong to them. Threads scan the CA bit vector and determine valid KV page columns, then scan the LA bit vector and determine local tile indices (both offset to global coordinates). Threads then cooperatively write these indices to their designated offset in \texttt{sparse\_n\_indices}. 

The decoding kernel overhead is 73\,$\mu$s, which is two orders of magnitude smaller than the per-layer sparse prefill compute (32\,ms) for MiniMax-M2.5 at 64K context. Therefore, decoding bit vectors adds minimal exposed overhead to the sparse prefill critical path.

\subsection{Custom Attention Kernel}
\label{sec:sparse-fa3}

A custom attention kernel extends the standard FlashAttention-3 Hopper kernel~\cite{fa3} with minimal modifications in order to enable sparse attention. The mainloop (TMA + warpgroup MMA) of standard dense N-block iteration is replaced by sparse index-driven iteration. At each iteration step $i$, the N-block index is read as $\texttt{sparse\_n\_indices}[i]$ instead of the dense $n_\text{max} - 1 - i$. This index drives the TMA descriptor for K/V tile loads.

The \texttt{sparse\_n\_indices} metadata resides in global memory and is accessed via scalar loads that hit L2 cache. The producer warpgroup (which issues TMA loads for K/V tiles) reads one KV index per N-block iteration to determine which tile to fetch and the consumer warpgroup (which performs WGMMA) then proceeds with selective recompute for that tile. The TMA pipeline for K/V tiles, the softmax accumulation, and the epilogue are left unmodified.

\begin{figure*}[htb!]
    \centering
    \includegraphics[width=\textwidth]{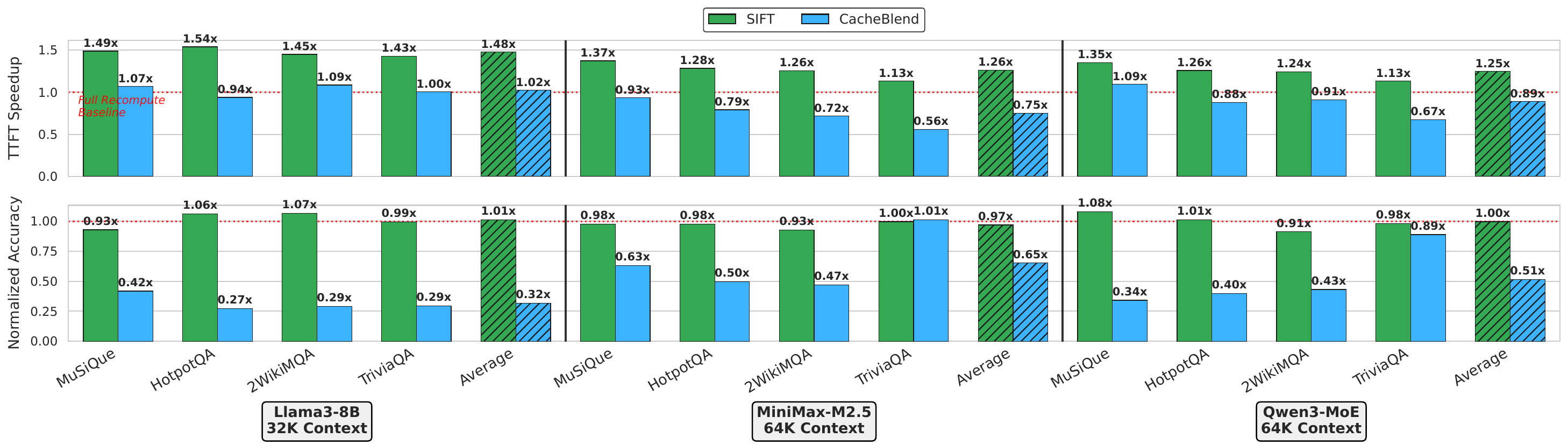}
    \caption{TTFT-speedup and accuracy of \Shortname and CacheBlend compared to full recompute on an 8x H200 system for LLama 8B (TP=1), Qwen3-235B-A22B (TP=8), and MiniMax M2.5 (TP=4): \Shortname gives consistent speedups while maintaining accuracy, while CacheBlend is bottlenecked by disk transfers and degrades accuracy.}
    \label{fig:all-models-benchmark}
\end{figure*}

\section{Evaluation Methodology}
\label{sec:eval-methodology}

\parhead{Serving Framework.}
We evaluate \Shortname by integrating it with vLLM v0.18.0~\cite{vllm} and LMCache~\cite{lmcache}. Our modifications extend LMCache's prefill framework to support \shortname's bit vector consumption and selective recompute of prefill, while preserving all other serving logic.

\parhead{Models.}
We evaluate on three models spanning a range of sizes and architectures: Llama-3.1-8B~\cite{llama3}, MiniMax-M2.5~\cite{minimax} and Qwen3-235B-A22B~\cite{qwen3}. All models are evaluated in BF16 precision. MiniMax-M2.5 natively trained MoE weights in FP8 precision so the MoE layers were computed in FP8.


\parhead{Datasets.}
We use LongBench~\cite{longbench} as our evaluation dataset and report accuracy using their own task-specific metrics. We evaluate on 4 datasets within LongBench: 2WikiMQA, HotpotQA, TriviaQA and Musique, each having 200 sample queries. To study long context behavior, we concatenate additional retrieved documents. All datasets, except TriviaQA, pertain to multi-document question answering tasks, specifically requiring the model to cross-attend to multiple documents to generate the appropriate answer.

\parhead{Baselines.}
We compare against two baselines: (i)~\emph{Full Recompute}, vLLM's default dense prefill with no caching, and (ii)~\emph{CacheBlend}~\cite{blend} to represent KV reuse with selective recompute. We used LMCache's default implementation of CacheBlend. 

\parhead{Metrics.}
We measure Time-to-First-Token~(TTFT) as our primary performance metric, since \Shortname optimizes the RAG prefill phase. LongBench uses F1 scores for all datasets. We report F1 scores normalized against full recompute and TTFT speedup over full recompute on identical samples and settings.

\parhead{Hardware.}
All experiments run on a single node with 8$\times$ NVIDIA H200 SXM 141\,GB GPUs~\cite{h200}. The host is equipped with 2\,TB CPU DRAM and 8$\times$ Micron 7450 3.84\,TB NVMe SSDs~\cite{micron-ssd} configured in RAID-0, providing 54.4\,GB/s aggregate disk bandwidth, and thus a max per-GPU share of 6.8 GB/s. CPU$\rightarrow$GPU transfers use PCIe Gen~5 $\times$16, yielding $\approx$63\,GB/s dedicated peak bandwidth.

Due to their significant size differences, we assume KV tensors of RAG documents are stored on SSDs and \Shortname metadata is stored on CPU DRAM. We also experimented with storing \Shortname metadata on SSD and observed that \shortname's disk reads incur negligible overhead due to its small size.







\section{Results}
\label{sec:evaluation}
In this section, we evaluate \Shortname and CacheBlend against Full Recompute. We analyze the impact on Time to First Token (TTFT) latency and accuracy of all three methods across different model configurations and context lengths.

\subsection{Impact on TTFT and Accuracy}
\label{sec:ttft-accuracy}

\parhead{\Shortname Performance.} \Shortname consistently achieves better TTFT than full recompute by reducing the compute during RAG prefill. The speedup provided by \Shortname varies and depends on the model and system configuration, context length and the unique sparsity ratio that \Shortname selects for each (document, attention head, layer) combination. As shown in \Cref{fig:all-models-benchmark}, \Shortname provides a speedup of $1.43\times$ to $1.54\times$ at 32K context lengths for dense Llama 8B model with the most accuracy degradation of 7\% on Musique and practically no degradation on the other LongBench tasks. This trend is followed for the larger MoE models too: MiniMax M2.5 has speedups upto 1.37x and Qwen3-235B-A22B upto 1.35x at 64K contexts, both with minimal accuracy degradation. Overall, the average accuracy degradation across all 12 datapoints is 1\%.

\parhead{KV Reuse Performance.} While CacheBlend does significantly less compute than both \Shortname and full recompute, it is actually \emph{slower} than both at certain context lengths because it becomes disk I/O bound even after pipelining the KV disk reads of the next layer with prefill computation of the current layer. Moreover, its 20\% recomputation is insufficient to maintain competitive accuracy with full recompute, and it suffers a massive 68.2\% accuracy degradation for Llama 8B at 32K context (\Cref{fig:all-models-benchmark}). TriviaQA is a single-hop question answering dataset and therefore does not require cross-document reasoning. Thus, it is the only dataset where CacheBlend maintains the accuracy of full recompute. 

\subsection{Impact of Context Length on TTFT and Accuracy}
\begin{figure}[htb!]
    \centering
    \includegraphics[width=\linewidth]{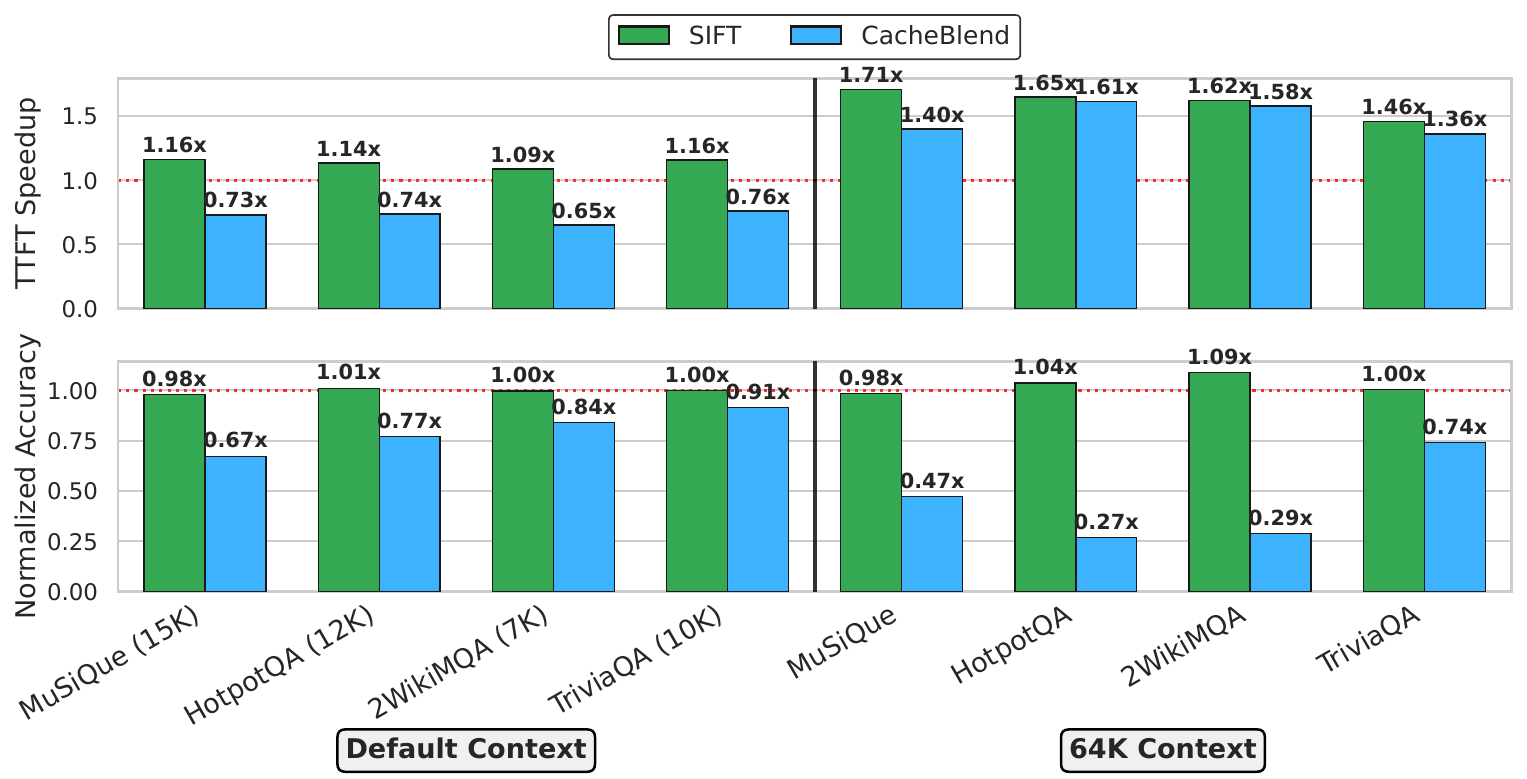}
    \caption{TTFT-Speedup and Accuracy on LLama3 8B (H200, TP=1) at 64K context length: At longer context length, CacheBlend's accuracy degrades while \Shortname remains similar.}
    \label{fig:llama3-benchmark}
    \vspace{-1em}
\end{figure}

The speedups of \Shortname increase as context lengths increase. Attention becomes more costly at longer contexts but also more sparse, allowing \Shortname to deliver higher speedups by skipping more attention tiles. \Cref{fig:llama3-benchmark} shows this increase in speedup as context lengths go from $15K$ to $64K$, with maximum speedup of 1.71x over full recompute.

For Cacheblend, as context lengths increase it becomes more GPU compute bound, and disk I/O can be either fully or partially hidden behind more GPU compute. Therefore, at 64K context for LLama3 8B, even CacheBlend's slow disk reads become faster than fully recomputing KVs, however its coarse-grained recomputation strategy still heavily degrades accuracy: \Cref{fig:llama3-benchmark} shows that as context lengths increase from  7K to 64K, accuracy degradation worsens from $20.25\%$ to $55.75\%$, whereas \Shortname maintains accuracy consistently at all context lengths.

\subsection{Storage Scaling of \Shortname}
\label{sec:storage-scaling}

The storage size for \Shortname is $O(P^2)$ bits, where $P$ is the number of tiled columns in the longest supported RAG document (as derived in \Cref{sec:ic-storage}). 

Notably, \shortname's storage requirement scales linearly with the number of documents and quadratically with the length of each document. KV Cache scales linearly with both number of documents and length of each document. However, since \shortname's size for a single token is in the order of bits and KV Cache size for a single token is in the order of KBs, \Shortname still remains extremely small in comparison.

For Llama 8B and typical RAG dataset document lengths, \shortname's size is about 24,000x smaller than KV Cache as shown in~\Cref{tab:cache-size}. Their sizes become equal only if the context length of each document was $33.6M$ tokens, which is far beyond typical RAG document lengths or even prompt lengths that we encounter in practice.


\begin{table}[t]
\centering
\caption{Storage scaling for \Shortname vs. KV Cache for different model architectures: \shortname's location metadata is extremely small compared to massive KV data.}
\label{tab:cache-size}
\footnotesize
\setlength{\tabcolsep}{4pt}
\begin{tabular}{@{}llrrc@{}}
\toprule
\textbf{Model} & \textbf{Ctx Length} & \textbf{KV Cache} & \textbf{\shortname} & \textbf{Reduction} \\
\midrule
\multirow{3}{*}{Llama-3.1-8B}
 & 4K   & 512\,MB  & 22\,KB  & \multirow{3}{*}{$23{,}831\times$} \\
 & 32K  & 4.0\,GB  & 176\,KB & \\
 & 125K & 15.6\,GB & 687\,KB & \\
\midrule
\multirow{3}{*}{Qwen3-MoE-22B}
 & 4K   & 752\,MB  & 129\,KB & \multirow{3}{*}{$5{,}958\times$} \\
 & 32K  & 5.9\,GB  & 1.0\,MB & \\
 & 125K & 22.9\,GB & 3.9\,MB & \\
\midrule
\multirow{3}{*}{MiniMax-M2.5}
 & 4K   & 992\,MB  & 64\,KB  & \multirow{3}{*}{$15{,}888\times$} \\
 & 32K  & 7.8\,GB  & 512\,KB & \\
 & 125K & 30.3\,GB & 2.0\,MB & \\
\bottomrule
\end{tabular}
\end{table}

\subsection{TTFT Breakdown}

\begin{figure}[htb!]
    \centering
    \includegraphics[width=\linewidth]{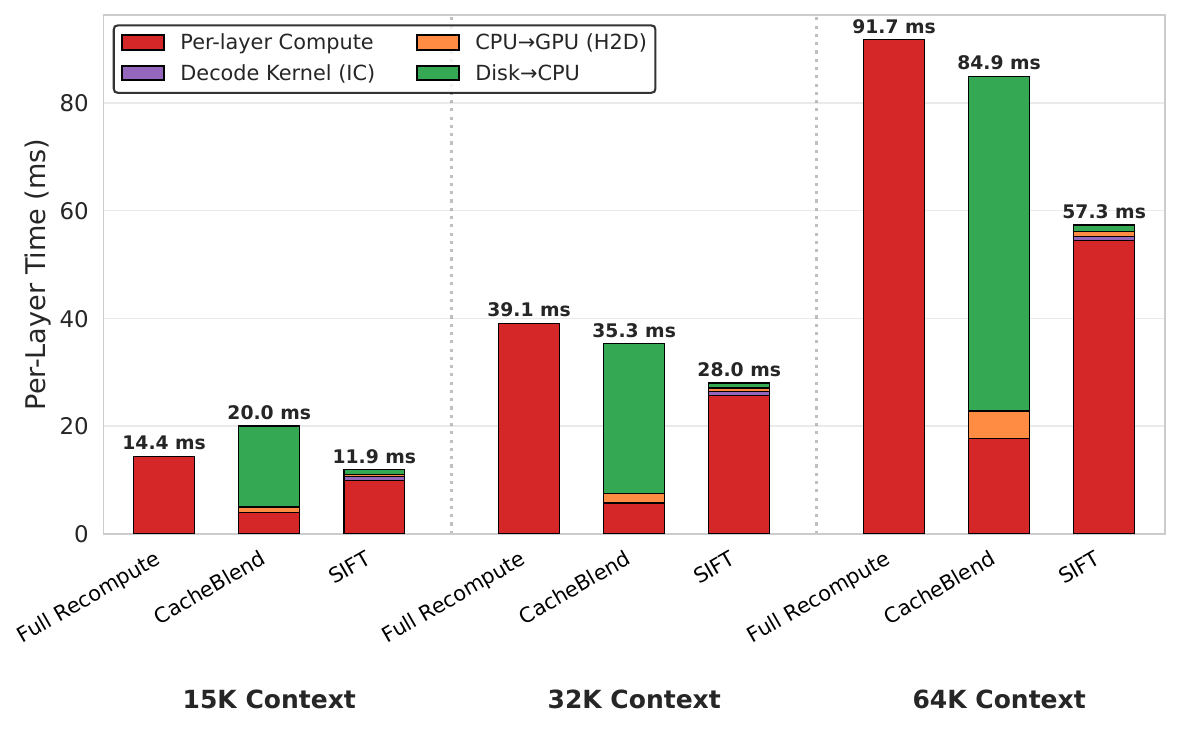}
    \caption{Breakdown of Disk$\rightarrow$CPU, CPU$\rightarrow$GPU data transfer time and compute time for Full Recompute, CacheBlend and \Shortname for Llama 8B: CacheBlend has a high overhead due to long-latency KV transfers while \shortname's metadata transfer incurs negligible overhead.}
    \label{fig:actual-ttft-breakdown}
\end{figure}

\Cref{fig:actual-ttft-breakdown} decomposes per-layer prefill into compute and data transfer time for all three modes for Llama 8B. For CacheBlend we read KV Cache of size $59.8$, $131.2$, and $235.5$\,MB per layer at 15K, 32K, and 64K context. CacheBlend's effective SSD read BW is only about $3.8$\,GB/s since it reads non-contiguous document KVs from disk. It's measured H2D BW is approximately $47$\,GB/s, which is close to peak for large MBs of transfer. The transfer latency breakdown for CacheBlend in \Cref{fig:actual-ttft-breakdown} depicts the exposed memory transfer latency that is not hidden even after pipelining KV transfer of layer $L+1$ from disk with GPU compute of layer $L$.

\Shortname metadata size is only $1.0$, $2.5$ and $4.5$\,MB across all layers at the same context lengths, so we read all layers from disk at once into CPU DRAM and then into GPU HBM. Thus, the disk$\rightarrow$CPU and CPU$\rightarrow$GPU transfer latency for \Shortname in \Cref{fig:actual-ttft-breakdown} represents fully exposed memory transfer time, not pipelined with GPU compute. Even then, the disk$\rightarrow$GPU transfer time takes up less than ${<}0.11\%$ of TTFT. For 15K, 32K and 64K contexts, the decode kernel in \Shortname takes about $0.04$, $0.075$ and $0.14$ ms for decoding the bit vectors of a given layer respectively. Therefore, the decode kernel has negligible overhead over per-layer compute time, contributing ${<}1\%$ to the overall TTFT.


\subsection{\Shortname Hyperparameter Analysis}

\begin{figure}[htb!]
    \centering
    \includegraphics[width=\linewidth]{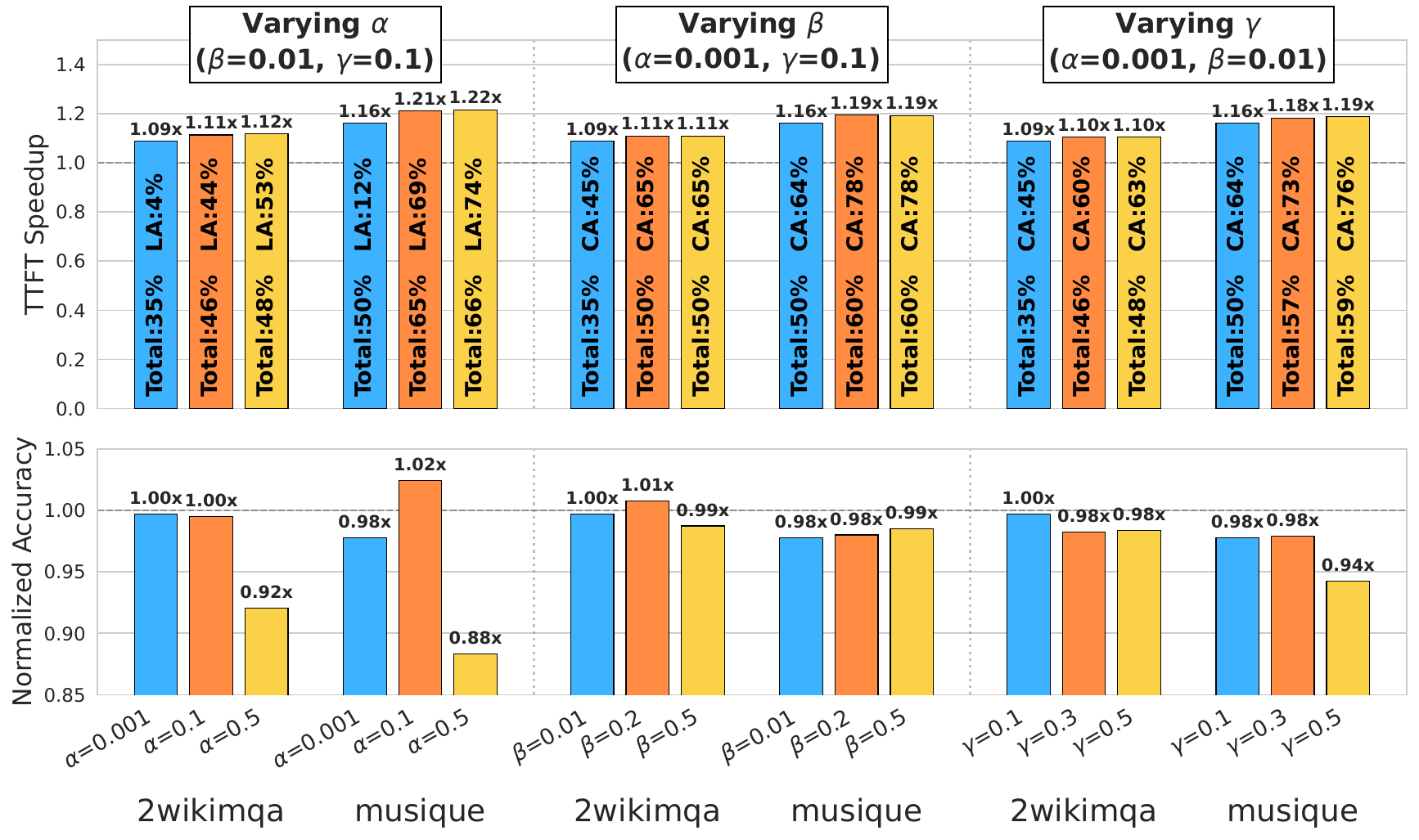}
    \caption{\shortname's TTFT and accuracy for varying hyperparameters for Llama 8B at $15K$ context. \textit{Total (\%)} is total sparsity, $LA(\%)$ is local-attention sparsity and $CA(\%)$ is cross-attention sparsity: Smaller values preserve accuracy, while large values have higher TTFT-speedups but lower accuracy.}
    \label{fig:hyperparam}
\end{figure}

\Cref{fig:hyperparam} shows a sensitivity analysis of \shortname's hyperparameters. Recall that \Shortname uses $\alpha$ to control local attention sparsity and $\beta$ and $\gamma$ to control cross-attention sparsity.  Increasing $\alpha$ beyond $0.1$ has diminishing returns in sparsity as attention becomes heavily sparse with only a few $(q, k)$ cells having high scores. This observation extends to $\beta$ and $\gamma$ too, where sparsity plateaus when $\beta$ and $\gamma$ are increased beyond $0.2$ and $0.3$, respectively.
Increasing the 3 hyperparameters leads to more sparsity overall, which helps improve TTFT but degrades accuracy. As maintaining accuracy is a key requirement in production scenarios, we use conservative values for the all 3 hyperparameters ($\alpha{=}0.001$, $\beta{=}0.01$ and $\gamma{=}0.1$). 




\subsection{Diverse Attention Patterns with \Shortname}

\begin{figure}[htb!]
\centering
\includegraphics[width=\linewidth]{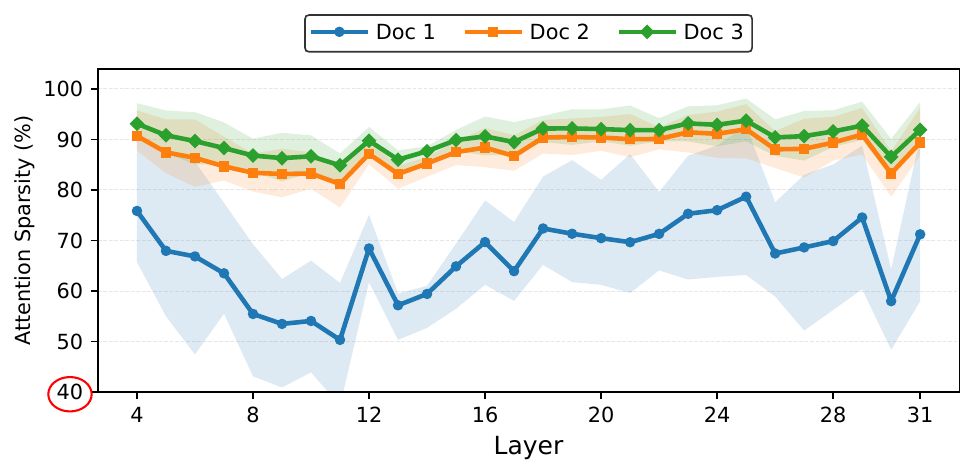}
\vspace{-0.15 in}
\caption{The sparsity pattern of \shortname \  across layers and heads for Llama 8B: Unlike CacheBlend, \Shortname selects varying number of recompute tiles for every (document, head, layer).}
\label{fig:diverse-sparsity}
\end{figure}

\Cref{fig:diverse-sparsity} depicts how \Shortname chooses diverse attention patterns for every (document, head, layer). The first document has lower attention sparsity than the rest because it only attends to itself. \Cref{fig:diverse-sparsity} also depicts the inter-quartile range (IQR, shaded band) of sparsity across heads. Even within a (document, layer), different heads recompute different number of tiles: IQR widths reach $10\%$ to $20\%$ for the first document and remain visible even for the more sparse second and third documents. Thus, \Shortname is able to recover diverse attention patterns at runtime.


\subsection{Energy Efficiency}
\label{sec:eval-energy}

SIFT not only improves performance but also energy efficiency compared to full recompute (SIFT reduces computation) and CacheBlend (SIFT reduces disk energy usage). We measure the energy efficiency of full recompute, \Shortname and CacheBlend for Qwen3-235B-A22B. GPU energy is integrated from NVML power samples, CPU DRAM energy is read from Intel RAPL counters and SSD energy is estimated using the Micron~7450~PRO datasheet~\cite{micron-ssd} (active read $11.5$\,W/drive and idle $5.0$\,W/drive).

Table~\ref{tab:energy-qwen3} reports TTFT, total energy $E$, performance per watt, the energy-delay product (EDP), and the energy-delay-squared product (ED$^2$P, metric often used for determining energy-efficiency of servers). \Shortname consistently delivers the best EDP and ED$^2$P. Its TTFT is $1.67\times$ faster than full recompute and $2.1\times$ faster than CacheBlend, giving it an ED$^2$P that is $3.1\times$ better than full recompute and $3.2\times$ better than CacheBlend. CacheBlend has the lowest absolute energy due to the lowest amount of GPU compute and therefore wins Perf/W, but its I/O-bound latency results in a worse EDP. 

\begin{table}[htb]
\centering
\footnotesize
\setlength{\tabcolsep}{3pt}
\caption{Perf/W, EDP and ED$^2$P for full recompute, \Shortname and CacheBlend: \Shortname gives best EDP and ED$^2$P.}
\label{tab:energy-qwen3}
\begin{tabular}{@{}llrrrrr@{}}
\toprule
Context & Mode & TTFT & $E$ & P/W & EDP & ED$^2$P \\
Length    &      & (s)  & (kJ)& ($\tfrac{1}{\text{kJ}}$) & (kJ$\cdot$s) & (kJ$\cdot$s$^{2}$) \\
\midrule
\multirow{3}{*}{32K}
 & Recompute     & 1.67 &  8.82 & 0.113 & 14.70 & 24.47 \\
 & \Shortname    & \textbf{1.00} &  7.87 & 0.127 & \textbf{7.83} & \textbf{7.80} \\
 & CacheBlend    & 2.13 & \textbf{5.51} & \textbf{0.181} & 11.72 & 24.90 \\
\midrule
\multirow{3}{*}{64K}
 & Recompute     & 2.92 & 17.94 & 0.056 & 52.31 & 152.53 \\
 & \Shortname    & \textbf{2.00} & \textbf{13.76} & \textbf{0.073} & \textbf{27.47} & \textbf{54.86} \\
 & CacheBlend    & 4.30 & 10.75 & 0.093 & 46.20 & 198.50 \\
\bottomrule

\end{tabular}
\vspace{-0.15 in}
\end{table}

\section{Related Works}
\label{sec:related-works}

\subsection{RAG Prefill Acceleration}
Prior works on RAG prefill acceleration store and reuse KV tensors of RAG documents in order to reduce compute and thus suffer from a prohibitive memory footprint and disk-bound data transfers (\Cref{sec:tradeoff}).

\parhead{Naive KV Reuse.} RAGCache~\cite{ragcache} employs prefix caching and tries to increase the KV cache hitrate across multiple queries by storing KVs in a prefix tree format. However, this limits its use to only the prefix portion of the prompts. PromptCache~\cite{promptcache} naively reuses precomputed KV without any recomputation, achieving the largest compute savings but the worst accuracy due to ignored cross-attention. 

\parhead{KV Reuse with Selective Recompute.} EPIC~\cite{epic} improves TTFT by deterministically recomputing only the first 64 tokens of every document in order to account for attention sink effects~\cite{attnsinks}. However, like CacheBlend, this coarse-grained recompute strategy does not account for diverse attention patterns. FusionRAG~\cite{fusionrag} makes documents cross-attend to each other offline, by leveraging the observation that documents retrieved by a similarity search to the user's query are also likely to be similar to each other. However, in addition to similarity-based retrieval, modern RAG retrieval search also incorporates diversity-aware re-rankers~\cite{mmr, opensearch-mmr,diverse-ranking, dartboard} which explicitly retrieves documents that are dissimilar to each other in order to increase information gain. Therefore, the assumption that retrieved documents are semantically similar to each other does not always hold, especially in regimes where RAG retrieval is required to be complex and diverse. \Shortname makes no assumption about the retrieval policy. 

\parhead{Finetuning Approaches.} TurboRAG~\cite{turborag} and KVLink~\cite{kvlink} finetune LLMs to improve accuracy of re-using KV tensors of RAG documents during an online query. However, fine-tuning incurs high computational costs and must be performed for each new LLM. With \shortname, we can run a one-time prefill for the most frequently accessed document clusters and easily amortize this cost over multiple queries.

\subsection{RAG Retrieval Acceleration}
Prior works~\cite{telerag, vectorliterag} accelerate the retrieval phase of the RAG pipeline, which performs similarity search over the RAG vector database containing millions of document embeddings. As RAG vector databases grow to tens of thousands of GB, their indices cannot fit in GPU memory, exposing CPU$\rightarrow$GPU transfer of cluster data into the critical path.

To accelerate retrieval latency, TeleRAG~\cite{telerag} observes that a user's initial query and its LLM-refined version produced during the pre-retrieval generation stage retrieve largely overlapping IVF clusters. It exploits this overlap to concurrently prefetch clusters from CPU to GPU during pre-retrieval LLM generation, hiding the transfer latency. VectorLiteRAG~\cite{vectorliterag} analytically partitions the IVF index between CPU and GPU based on access skew and SLO targets, allocating hot clusters to GPU and cold clusters to CPU. PipeRAG~\cite{piperag} targets iterative RAG pipelines in which retrieval occurs periodically during generation, pipelining each retrieval with the concurrent decode stage. These works are orthogonal to \shortname. They reduce \emph{retrieval} latency, while \Shortname reduces \emph{prefill} latency. A RAG system can deploy both.

\subsection{Sparse Prefill Attention}
Prior works that exploit sparsity in prefill attention can be broadly delineated into two types: (1) Static sparse attention techniques~\cite{streamingllm,minference,attnsinks} that identify structured coarse-grained attention patterns offline and (2) Dynamic sparse attention techniques~\cite{FlexPrefill,sampleattn} that determine sparsity patterns at runtime. Runtime determinism enables a context-informed sparsity pattern that achieves better accuracy than static patterns, but also incurs non-trivial runtime overheads. Ideally, we would want the fine-grained sparsity pattern found during runtime, but with the zero overhead of static sparse attention. 

FlexPrefill~\cite{FlexPrefill} is one such dynamic sparsity approach that determines attention patterns on-the-fly by computing a representative attention map from the last block of queries against all keys. While FlexPrefill achieves impressive TTFT reduction at hyper-long contexts, its on-the-fly sparsity prediction incurs significant overhead even at moderate context length (e.g. runtime overhead of ~$50\%$ of sparse attention at 32K context). It provides meaningful speedups only at contexts >128K.


Both static and dynamic sparsity techniques have so far been \emph{RAG-agnostic}: they treat every prefill as a fresh, independent computation and make no use of the fact that RAG contexts are known apriori. Whereas \Shortname exploits the context reuse property of RAG workloads to enable fine-grained context-dependent sparsity that is determined offline and incurs negligible runtime overheads.

\section{Conclusion}
Retrieval-Augmented Generation (RAG) prepends relevant documents to the user query, increasing context length and leading to a longer time to first token (TTFT). Prior work precomputes KVs for RAG documents offline and reuses them during online queries. However, it suffers from two major limitations: (1) Accuracy degradation due to coarse-grained online recomputation that does not account for the unique attention patterns of every (document, head, layer) and (2) Limited speedups because of expensive transfers of KVs from storage to compute. We present \textit{\fullname}, which identifies and exploits the attention invariance properties of RAG documents to identify and selectively recompute high attention scores. The metadata of \Shortname is ~24,000$\times$ smaller than KV cache and delivers up to a 1.71$\times$ TTFT speedup over full recompute, while maintaining an average accuracy of within 1\%.  
\label{sec:conclusion}

\clearpage



\bibliographystyle{ACM-Reference-Format}
\bibliography{refs}

\end{document}